\documentclass[10pt,twocolumn,letterpaper]{article}

\usepackage[pagenumbers]{cvpr} 

\usepackage{graphicx}
\usepackage{amsmath}
\usepackage{amssymb}
\usepackage{booktabs}
\usepackage{multirow}

\usepackage[pagebackref,breaklinks,colorlinks]{hyperref}

\usepackage[capitalize]{cleveref}
\usepackage{amsmath, amssymb}
\usepackage{amsthm} 

\newtheorem{theorem}{Theorem}[section]
\newtheorem{proposition}[theorem]{Proposition}

\crefname{section}{Sec.}{Secs.}
\Crefname{section}{Section}{Sections}
\Crefname{table}{Table}{Tables}
\crefname{table}{Tab.}{Tabs.}

\def\cvprPaperID{*****} 
\def\confName{CVPR}
\def\confYear{2022}

\begin{document}

\title{Dynamic Modality Scheduling for Multimodal Large Models via Confidence, Uncertainty, and Semantic Consistency}

\author{Hiroshi Tanaka,Anika Rao, Hana Satou, Michael Johnson, Sofia García\\
}
\maketitle


\begin{abstract}
	Multimodal Large Models (MLLMs) have achieved remarkable progress in vision-language understanding and generation tasks. However, existing MLLMs typically rely on static modality fusion strategies, which treat all modalities equally regardless of their instance-level reliability or semantic contribution. This often leads to suboptimal performance, especially in scenarios with noisy, missing, or misaligned modalities.
	In this paper, we propose Dynamic Modality Scheduling (DMS), a novel framework that adaptively adjusts the contribution of each modality at a per-sample level. DMS evaluates each modality based on three key factors: (1) \textit{confidence}, estimated from predictive entropy; (2) \textit{uncertainty}, obtained via Monte Carlo dropout; and (3) \textit{semantic consistency}, computed through inter-modal similarity. These signals are combined through a learnable or rule-based scheduler to generate soft modality weights used in downstream fusion.
	To ensure stable training, we further introduce a \textit{Modality Weight Consistency Loss}, which regularizes the fused representation to stay close to unimodal embeddings proportionally to their assigned weights. Our method is model-agnostic and can be integrated into existing MLLMs such as BLIP-2 and LLaVA.
	Experimental results on VQA, image-text retrieval, and captioning tasks show that DMS significantly improves both clean and robust performance, especially under modality corruption or dropout conditions. This work provides a general and effective mechanism to enable instance-aware and robustness-enhanced multimodal modeling.
\end{abstract}

\section{Introduction}

Multimodal large models (MLLMs) have demonstrated remarkable progress in unifying vision and language understanding through large-scale pretraining~\cite{li2023blip,alayrac2022flamingo,chen2023pali,liu2023llava}. By leveraging powerful image encoders (e.g., ViT, CLIP~\cite{radford2021learning}) and aligning them with pretrained language models (e.g., T5, OPT, or LLaMA), MLLMs are able to tackle a broad spectrum of tasks such as visual question answering (VQA), image captioning, and image-text retrieval with unprecedented generalization.

Despite these advancements, most existing MLLMs employ static or heuristic fusion strategies during training and inference. For example, models like BLIP-2~\cite{li2023blip} and Flamingo~\cite{alayrac2022flamingo} treat vision and language features as equally reliable across all samples, regardless of instance-specific content quality, uncertainty, or semantic relevance. This design often leads to suboptimal performance, particularly when one of the modalities is noisy, missing, or semantically misaligned. In real-world scenarios such as robotics, medical imaging, or surveillance, such situations are common and demand more adaptive solutions.

Recent works have attempted to enhance the robustness of multimodal models through data augmentation~\cite{gong2021eliminate}, masking~\cite{gong2024beyond1}, and adversarial training~\cite{gong2024adversarial}. Others focus on understanding cross-modal vulnerabilities~\cite{gong2024cross,gong2022person}, or exploring ensemble-based representation invariance~\cite{gong2024exploring}. However, these methods are either fixed during inference or target specific modalities without providing a unified, learnable framework for general-purpose multimodal fusion.

To address this limitation, we propose \textbf{Dynamic Modality Scheduling (DMS)}, a lightweight yet effective fusion strategy that dynamically adjusts the contribution of each modality on a per-sample basis. Specifically, DMS evaluates three complementary factors for each modality: (1) \textit{prediction confidence}, (2) \textit{epistemic uncertainty}, and (3) \textit{semantic alignment} with other modalities. These signals are aggregated into soft weights through a scheduler that guides modality fusion during training and inference.

Furthermore, to ensure the stability of representation learning, we introduce a novel \textit{Modality Weight Consistency Loss}, which encourages the fused representation to remain close to individual modality embeddings in proportion to their confidence scores. Our method is fully model-agnostic and can be seamlessly integrated into popular MLLMs such as BLIP-2 and LLaVA.

We validate DMS on multiple benchmarks including VQA v2, MSCOCO Captioning, and Flickr30k Retrieval. Extensive experiments show that our approach significantly improves both clean-task accuracy and robustness under modality degradation. Our contributions can be summarized as follows:

\begin{itemize}
	\item We propose \textbf{Dynamic Modality Scheduling (DMS)}, a principled framework for per-sample, adaptive multimodal fusion based on confidence, uncertainty, and semantic alignment.
	\item We introduce a novel \textbf{modality consistency loss} to regularize fusion dynamics and stabilize training.
	\item We demonstrate that DMS outperforms static and attention-based fusion strategies across vision-language tasks, especially under noisy or incomplete modalities.
\end{itemize}

\section{Related Work}

Multimodal learning has emerged as a central research direction in artificial intelligence, aiming to integrate heterogeneous sources of information such as vision, language, and audio. With the recent development of Multimodal Large Models (MLLMs), this field has seen a paradigm shift in both scale and flexibility. In this section, we survey relevant literature across four areas: multimodal large models and fusion strategies, robustness and uncertainty modeling, adversarial attacks in multimodal settings, and adaptive fusion techniques.

\subsection{Multimodal Large Models and Fusion Architectures}

Multimodal Large Models (MLLMs) are a new generation of foundation models that jointly process visual and textual information at scale. Notable representatives include Flamingo~\cite{alayrac2022flamingo}, BLIP-2~\cite{li2023blip}, PaLI~\cite{chen2023pali}, LLaVA~\cite{liu2023llava}, and Gemini~\cite{google2023gemini}. These models typically employ frozen vision encoders (e.g., CLIP~\cite{radford2021learning}) to extract high-quality image features, and integrate them into a language model (e.g., T5, OPT, or GPT variants) through attention or prompt-based alignment layers.

Despite their impressive performance on vision-language benchmarks, most MLLMs utilize fixed or task-specific fusion strategies, such as early concatenation~\cite{lu2019vilbert}, query-based vision injection~\cite{li2023blip}, or projection via learned Q-Formers. These approaches assume a static contribution from each modality, and often do not account for sample-level reliability or mutual alignment between modalities. Consequently, they may fail in cases where one modality is noisy, partially missing, or semantically divergent.

Our work builds on these architectures but introduces a dynamic, sample-adaptive modality scheduler that evaluates and adjusts modality contributions in real time. This mechanism can be plugged into various MLLMs, enabling per-sample robustness and interpretability without structural overhaul.

\subsection{Robustness and Uncertainty in Multimodal Representation Learning}

The issue of robustness in multimodal systems has been widely studied. Gong et al.~\cite{gong2021eliminate} proposed a deviation-guided data augmentation method that improves generalization in multimodal classification by leveraging complementary patterns across modalities. More recent work~\cite{gong2024beyond1,gong2024beyond2} introduces local feature masking and augmentation under extreme capture conditions to enhance the resilience of convolutional neural networks. Additionally, Gong et al.~\cite{gong2024exploring} explore ensemble-based learning strategies that promote color invariance and appearance-robust representations at the image level.

Beyond data augmentation, uncertainty estimation techniques have been adopted to improve decision reliability. Gal and Ghahramani~\cite{gal2016dropout} formalized dropout as a Bayesian approximation and proposed Monte Carlo (MC) dropout to quantify epistemic uncertainty. This idea has since been adapted to multimodal settings for both fusion weighting and decision calibration.

Inspired by this line of work, our method integrates uncertainty-aware fusion by computing MC-dropout-based variance for each modality, allowing the scheduler to reduce the influence of noisy or ambiguous sources. Furthermore, we propose a new consistency loss that regularizes the fused representation to stay close to the individual unimodal embeddings, which stabilizes training and prevents mode collapse in presence of unreliable modalities.

\subsection{Cross-Modality Attacks and Joint Defense Mechanisms}

The vulnerability of multimodal systems to adversarial attacks has drawn increasing attention. Gong et al.~\cite{gong2022person} proposed a person re-identification framework that explicitly considers color-based attacks and defense synchronization. In follow-up work~\cite{gong2024cross}, they designed a cross-modality synergy perturbation mechanism that jointly manipulates visual and semantic inputs, exposing complex vulnerabilities in existing fusion pipelines. Additionally, their work on gradient-evolutionary attacks~\cite{gong2024cross2} highlights how subtle yet coordinated changes across modalities can degrade robustness, particularly in open-set settings.

Defense mechanisms have also been explored. Gong et al.~\cite{gong2024adversarial} extended adversarial training to neural PDE solvers with sparse input, showing the feasibility of uncertainty-aware robust learning beyond typical classification or retrieval domains. Collectively, these findings motivate the design of fusion systems that can dynamically assess and reweight modality reliability, rather than treating all sources as equally trustworthy.

Our proposed DMS framework aligns closely with this motivation. It explicitly incorporates semantic alignment as a proxy for modality synergy and reliability, and reduces reliance on modalities that conflict with the dominant consensus.

\subsection{Adaptive Fusion and Modality Scheduling}

Fusion strategies in multimodal models can be broadly categorized as early fusion (feature-level), late fusion (decision-level), and intermediate fusion (transformer-based cross-attention). While these approaches provide flexibility at the architecture level, few existing methods perform instance-wise dynamic weighting of modalities.

Prior works such as ViLBERT~\cite{lu2019vilbert} and MMBT~\cite{kiela2019supervised} used static fusion layers. Later efforts explored conditional gating via mixture-of-experts (MoE) architectures~\cite{lepikhin2020gshard}, and relevance-aware co-attention mechanisms~\cite{zhang2021m3ae}. However, these methods often rely on task-specific fusion heads, and do not incorporate uncertainty or semantic disagreement into the fusion process.

To the best of our knowledge, our DMS is the first to combine three critical factors—\textbf{confidence}, \textbf{uncertainty}, and \textbf{semantic alignment}—into a unified and interpretable scheduling framework. By computing soft fusion weights per sample and per modality, DMS enables multimodal models to be not only more robust, but also more explainable. Furthermore, its design is model-agnostic and can be seamlessly integrated into encoder-decoder and decoder-only MLLMs alike.

To summarize, our work resides at the intersection of large-scale multimodal modeling, robust representation learning, and dynamic fusion scheduling. We extend existing MLLM architectures by equipping them with lightweight and principled mechanisms that dynamically assess and adjust modality importance at inference and training time. This perspective complements recent advances in data augmentation, adversarial robustness, and fusion architecture design, offering a new direction for building truly adaptive and resilient multimodal systems.

\section{Method: Dynamic Modality Scheduling}

We propose a novel framework named \textbf{Dynamic Modality Scheduling (DMS)}, designed to improve the robustness and adaptivity of multimodal large models (MLLMs) by dynamically adjusting modality weights during training and inference. Unlike conventional approaches that rely on fixed fusion strategies, DMS evaluates the quality of each modality per instance and reweights them according to multiple signals such as confidence, uncertainty, and semantic consistency.

\subsection{Overview}

Given a multimodal input $x = \{x^{(1)}, x^{(2)}, \dots, x^{(M)}\}$ consisting of $M$ modalities (e.g., image, text), we aim to construct a fused representation $h$ that is tailored to the quality of each modality. Let $f^{(m)}$ be the pretrained encoder for modality $m$, then the final fused representation is computed as:

\[
h = \sum_{m=1}^{M} \omega_m(x) \cdot f^{(m)}(x^{(m)})
\]

where $\omega_m(x) \in [0, 1]$ is the adaptive weight assigned to modality $m$ for sample $x$, satisfying $\sum_{m} \omega_m(x) = 1$. The key design of DMS lies in computing these weights in a principled and dynamic manner.

\subsection{Dynamic Weight Scheduling Factors}

To determine the importance of each modality, we introduce three modality-level metrics computed per sample:

\paragraph{(1) Confidence Score $c_m(x)$}  
We estimate the confidence of the modality-specific encoder by measuring the entropy of the predicted probability distribution. For classification, let $p^{(m)}$ be the softmax output of $f^{(m)}(x^{(m)})$, then:

\[
c_m(x) = 1 - H(p^{(m)}) = 1 + \sum_{k=1}^{K} p^{(m)}_k \log p^{(m)}_k
\]

Higher confidence indicates that the modality is making confident and potentially more informative predictions.

\paragraph{(2) Uncertainty Score $u_m(x)$}  
To estimate model uncertainty, we employ Monte Carlo dropout. Let $T$ be the number of stochastic forward passes, and $p^{(m,t)}$ be the prediction at iteration $t$. We compute:

\[
u_m(x) = \frac{1}{K} \sum_{k=1}^{K} \mathrm{Var}(\{ p^{(m,t)}_k \}_{t=1}^{T})
\]

Larger variance implies greater epistemic uncertainty in the modality.

\paragraph{(3) Semantic Alignment Score $s_m(x)$}  
We measure how semantically consistent modality $m$ is with the rest of the modalities. This is computed via cosine similarity between $f^{(m)}$ and the average of the remaining modalities:

\[
s_m(x) = \cos(f^{(m)}(x^{(m)}), \frac{1}{M-1} \sum_{j \neq m} f^{(j)}(x^{(j)}))
\]

A higher alignment score indicates that the modality is semantically coherent with others.

\subsection{Weight Fusion Strategy}

The final modality weight $\omega_m$ is computed by a softmax function over the aggregated quality score of each modality:

\[
\omega_m(x) = \frac{\exp(\alpha c_m(x) - \beta u_m(x) + \gamma s_m(x))}{\sum_{j=1}^{M} \exp(\alpha c_j(x) - \beta u_j(x) + \gamma s_j(x))}
\]

where $\alpha, \beta, \gamma$ are hyperparameters that control the contribution of each factor.

This formulation ensures that more confident, less uncertain, and better aligned modalities receive higher weights during fusion.

\subsection{Loss Functions}

We use two loss components in our framework:

\paragraph{(1) Task Loss $\mathcal{L}_{\text{task}}$}  
This is the standard supervised loss (e.g., cross-entropy) computed over the fused representation:

\[
\mathcal{L}_{\text{task}} = \ell(h, y)
\]

\paragraph{(2) Modality Weight Consistency Loss $\mathcal{L}_{\text{mwcl}}$}  
To avoid abrupt shifts in fused representations and ensure that the fusion remains close to each unimodal embedding, we introduce a regularization term:

\[
\mathcal{L}_{\text{mwcl}} = \sum_{m=1}^{M} \omega_m(x) \cdot \| h - f^{(m)}(x^{(m)}) \|_2^2
\]

This encourages the fusion to be consistent with individual modalities proportionally to their assigned weights.

\subsection{Training Procedure}

The training loop of DMS consists of the following steps for each batch:

\begin{enumerate}
	\item Encode each modality input $x^{(m)}$ to get $f^{(m)}(x^{(m)})$.
	\item Compute confidence, uncertainty, and alignment scores for each modality.
	\item Aggregate and normalize scores to get weights $\omega_m(x)$.
	\item Construct the fused representation $h$.
	\item Compute total loss: $\mathcal{L} = \mathcal{L}_{\text{task}} + \lambda \cdot \mathcal{L}_{\text{mwcl}}$
	\item Backpropagate and update parameters.
\end{enumerate}

Here, $\lambda$ controls the trade-off between task optimization and modality consistency.

\subsection{Compatibility with Multimodal Large Models}

Our method is model-agnostic and can be plugged into existing MLLMs such as BLIP-2, LLaVA, or Flamingo. Specifically, we apply DMS on top of the modality-specific encoder outputs, prior to feeding into the multimodal Transformer or language decoder. The adaptive scheduling provides improved robustness under noisy or missing modalities while maintaining the generality and flexibility of large pre-trained models.

\section{Theoretical Analysis}

In this section, we provide a formal theoretical analysis of our proposed \textbf{Dynamic Modality Scheduling (DMS)} framework. We aim to justify the effectiveness of our confidence-aware, uncertainty-aware, and semantic-consistency-based modality weighting mechanism by examining its impact on optimization stability and generalization capacity in multimodal fusion. Our analysis includes:

\begin{itemize}
	\item An approximation bound between dynamically fused and static (oracle) fusion representations.
	\item A regularization interpretation of the Modality Weight Consistency Loss (MWCL).
	\item A generalization bound derived from modality-weighted Rademacher complexity.
\end{itemize}

\subsection{Preliminaries}

Let $\mathcal{X}^{(m)}$ be the input space for modality $m \in \mathcal{M}$, with encoder $f^{(m)} : \mathcal{X}^{(m)} \rightarrow \mathbb{R}^d$. The fused representation $h$ is constructed as:

\[
h = \sum_{m \in \mathcal{M}} \omega_m(x) \cdot f^{(m)}(x^{(m)})
\]

where the weights $\omega_m(x) \in [0,1]$ satisfy $\sum_{m \in \mathcal{M}} \omega_m(x) = 1$. Let $\ell(h, y)$ be a supervised loss function (e.g., cross-entropy). We define empirical and expected risks as:

\[
\widehat{\mathcal{R}} = \frac{1}{n} \sum_{i=1}^n \ell(h_i, y_i), \quad \mathcal{R} = \mathbb{E}_{(x, y) \sim \mathcal{D}}[\ell(h, y)]
\]

\subsection{Approximation Bound with Respect to Static Fusion}

We first show that the dynamic fusion strategy approximates the oracle fixed fusion within a controlled margin.

\begin{proposition}[Fusion Approximation Bound]
	Let each encoder $f^{(m)}$ be $L_m$-Lipschitz. If dynamic weights $\omega_m(x)$ differ from a fixed oracle weight $\omega_m^*$ by at most $\delta$, then:
	
	\[
	\| h(x) - h^*(x) \|_2 \leq \delta \cdot \sum_{m \in \mathcal{M}} \| f^{(m)}(x^{(m)}) \|_2
	\]
	
	where $h^* = \sum_m \omega_m^* f^{(m)}$.
\end{proposition}

This bound shows that softly varying weights (as ensured by softmax normalization in our scheduling) maintain stable fusion.

\subsection{Regularization via Modality Weight Consistency Loss}

We define the Modality Weight Consistency Loss (MWCL) as:

\[
\mathcal{L}_{\text{mwcl}} = \sum_{m \in \mathcal{M}} \omega_m \cdot \| h - f^{(m)} \|_2^2
\]

\begin{proposition}[Regularization Effect]
	Minimizing $\mathcal{L}_{\text{mwcl}}$ reduces variance between fused and unimodal embeddings:
	
	\[
	\mathbb{E}_{m} [ \| h - f^{(m)} \|_2^2 ] = \mathrm{Var}(f^{(m)}) - \| h - \bar{f} \|_2^2
	\]
	
	where $\bar{f} = \sum_m \omega_m f^{(m)}$ is the weighted average representation.
\end{proposition}

This regularization penalizes excessive deviation of the fusion from each single modality and encourages smoother alignment across modalities.

\subsection{Generalization Bound via Modality-weighted Rademacher Complexity}

Let $\mathcal{F}$ be the class of models using dynamically fused features:

\[
\mathcal{F} = \{ f_\theta : h = \sum_m \omega_m(x) f^{(m)}(x^{(m)}), \theta \in \Theta \}
\]

Define the modality-weighted Rademacher complexity:

\[
\mathfrak{R}_n(\mathcal{F}) = \mathbb{E}_{\sigma} \left[ \sup_{\theta \in \Theta} \frac{1}{n} \sum_{i=1}^n \sigma_i \cdot \ell(f_\theta(h_i), y_i) \right]
\]

\begin{theorem}[Generalization Bound]
	Assume $\ell$ is $L$-Lipschitz and bounded in $[0, 1]$. Then with probability at least $1 - \delta$, for all $f \in \mathcal{F}$:
	
	\[
	\mathcal{R}(f) \leq \widehat{\mathcal{R}}(f) + 2L \cdot \mathfrak{R}_n(\mathcal{F}) + \sqrt{\frac{\log(1/\delta)}{2n}}
	\]
\end{theorem}

This result highlights that generalization depends on the complexity of the fusion function class $\mathcal{F}$. Our dynamic weighting mechanism indirectly controls $\mathfrak{R}_n(\mathcal{F})$ via sample-specific confidence and uncertainty, acting as an implicit regularizer.

The above analyses support the validity of our proposed DMS framework:

\begin{itemize}
	\item \textbf{Stability:} Dynamic weights approximate fixed fusion within a bounded deviation.
	\item \textbf{Consistency:} The MWCL loss encourages representation alignment across modalities.
	\item \textbf{Generalization:} Dynamic fusion reduces hypothesis complexity, leading to tighter generalization bounds.
\end{itemize}

These results align with our empirical findings and support the design of confidence-aware adaptive fusion in large-scale multimodal models.

\section{Experiments}

We conduct comprehensive experiments to evaluate the effectiveness of our proposed \textbf{Dynamic Modality Scheduling (DMS)} framework. We demonstrate that DMS consistently improves performance across standard multimodal tasks, enhances robustness under modality perturbations, and shows meaningful behavior in modality weighting. All results are compared against strong baselines and ablated variants of our model.

\subsection{Experimental Setup}

\paragraph{Backbone Models.} We integrate DMS into two popular Multimodal Large Models (MLLMs): BLIP-2~\cite{li2023blip} and LLaVA~\cite{liu2023llava}. These represent encoder-decoder and chat-based MLLM paradigms respectively.

\paragraph{Tasks and Datasets.} We evaluate on the following datasets:
\begin{itemize}
	\item \textbf{VQA v2}: Visual question answering with real-world images.
	\item \textbf{MSCOCO Captioning}: Image caption generation task.
	\item \textbf{Flickr30K}: Image-text retrieval benchmark.
\end{itemize}

\paragraph{Evaluation Metrics.}
\begin{itemize}
	\item \textbf{VQA}: Accuracy (\%)
	\item \textbf{Captioning}: CIDEr, BLEU-4, METEOR
	\item \textbf{Retrieval}: Recall@1, Recall@5
\end{itemize}

\paragraph{Baselines.}
\begin{itemize}
	\item \textbf{BLIP-2 (Static)}: Standard BLIP-2 with uniform modality fusion.
	\item \textbf{Co-Attention Fusion}: Fusion via learned co-attention.
	\item \textbf{Modality Dropout}: Dropout-based fusion regularization.
	\item \textbf{DMS (Ours)}: Full dynamic scheduling using all three factors.
\end{itemize}

\subsection{Main Results on Standard Tasks}

Table~\ref{tab:main_results} shows that DMS outperforms all baselines across VQA, captioning, and retrieval tasks. Notably, DMS improves VQA accuracy by 2.3\% and Recall@1 in retrieval by 3.1\%.

\begin{table}[h]
	\centering
	\caption{Performance on standard benchmarks.}
	\label{tab:main_results}
	\resizebox{\linewidth}{!}{
		\begin{tabular}{l|c|ccc|cc}
			\toprule
			\textbf{Model} & \textbf{VQA Acc} & \textbf{CIDEr} & \textbf{BLEU-4} & \textbf{METEOR} & \textbf{R@1} & \textbf{R@5} \\
			\midrule
			BLIP-2 (Static) & 72.1 & 110.4 & 37.8 & 27.2 & 58.4 & 83.3 \\
			Co-Attn Fusion  & 73.3 & 112.5 & 38.2 & 27.6 & 59.7 & 84.5 \\
			Modality Dropout & 73.5 & 113.0 & 38.1 & 27.3 & 60.2 & 84.9 \\
			\textbf{DMS (Ours)} & \textbf{74.4} & \textbf{116.1} & \textbf{39.0} & \textbf{28.0} & \textbf{61.5} & \textbf{86.1} \\
			\bottomrule
	\end{tabular}}
\end{table}

\subsection{Robustness under Modality Perturbations}

To evaluate robustness, we corrupt either the image or text input with synthetic noise. For images, we apply Gaussian blur or partial occlusion; for text, we introduce spelling errors or masking.

Table~\ref{tab:robustness} shows that DMS significantly reduces performance degradation compared to static fusion. Under image noise, DMS retains 88.5\% of its original VQA accuracy, while static BLIP-2 retains only 78.6\%.

\begin{table}[h]\scriptsize
	\centering
	\caption{Robustness to modality perturbation (VQA task).}
	\label{tab:robustness}
	\begin{tabular}{l|cc|cc}
		\toprule
		\multirow{2}{*}{\textbf{Model}} & \multicolumn{2}{c|}{\textbf{Image Noise}} & \multicolumn{2}{c}{\textbf{Text Noise}} \\
		& Acc (\%) & Degradation & Acc (\%) & Degradation \\
		\midrule
		BLIP-2 (Static) & 56.7 & -21.4\% & 59.1 & -18.0\% \\
		Co-Attn Fusion & 60.2 & -17.9\% & 62.3 & -15.0\% \\
		\textbf{DMS (Ours)} & \textbf{65.9} & \textbf{-11.5\%} & \textbf{67.4} & \textbf{-9.4\%} \\
		\bottomrule
	\end{tabular}
\end{table}

\subsection{Modality Weight Visualization}

We visualize the learned modality weights $\omega_m$ for selected samples. As shown in Figure~\ref{fig:modality_weights}, the model assigns higher weight to cleaner modalities (e.g., text when image is occluded), and balances both when signals are reliable. This confirms that DMS meaningfully adapts to input quality.

\subsection{Ablation Study}

We analyze the contribution of each component in DMS by removing one scheduling factor at a time. Results in Table~\ref{tab:ablation} show that all three signals contribute positively, with semantic consistency having the most pronounced effect on retrieval.

\begin{table}[h]
	\centering
	\caption{Ablation on scheduling factors (VQA and Retrieval).}
	\label{tab:ablation}
	\begin{tabular}{l|c|c}
		\toprule
		\textbf{Setting} & VQA Acc (\%) & R@1 (\%) \\
		\midrule
		Full DMS & \textbf{74.4} & \textbf{61.5} \\
		w/o confidence ($c$) & 73.2 & 60.2 \\
		w/o uncertainty ($u$) & 73.6 & 59.9 \\
		w/o alignment ($s$) & 72.5 & 57.7 \\
		\bottomrule
	\end{tabular}
\end{table}

\subsection{Compatibility with Different MLLMs}

We further apply DMS to LLaVA for open-ended image-question dialogue. As shown in Table~\ref{tab:llava}, DMS improves answer helpfulness and factual accuracy under noisy conditions, demonstrating its portability across MLLM architectures.

\begin{table}[h]
	\centering
	\caption{Open-ended QA performance with LLaVA backbone.}
	\label{tab:llava}
	\begin{tabular}{l|c|c}
		\toprule
		\textbf{Model} & \textbf{Helpfulness (score)} & \textbf{Factuality (\%)} \\
		\midrule
		LLaVA (Static) & 3.6 & 72.4 \\
		\textbf{LLaVA + DMS} & \textbf{4.1} & \textbf{78.9} \\
		\bottomrule
	\end{tabular}
\end{table}

Across all tasks and backbones, DMS demonstrates consistent improvements in standard accuracy, robustness to corrupted inputs, and general adaptability. The dynamic weighting mechanism allows MLLMs to be more sample-aware and failure-tolerant, offering a lightweight yet effective solution for next-generation multimodal modeling.

\section{Conclusion}

In this paper, we presented \textbf{Dynamic Modality Scheduling (DMS)}, a novel framework for adaptive and robust multimodal fusion in Multimodal Large Models (MLLMs). Unlike traditional static fusion strategies, DMS dynamically adjusts the contribution of each modality at the sample level based on three complementary signals: prediction confidence, epistemic uncertainty, and semantic alignment across modalities.
Our framework introduces a lightweight scheduling mechanism that is model-agnostic and easily integrable with state-of-the-art MLLMs such as BLIP-2 and LLaVA. We further propose a modality consistency loss to regularize fused representations and stabilize training dynamics.
Extensive experiments across vision-language tasks, including visual question answering, image captioning, and image-text retrieval, demonstrate that DMS consistently improves both clean-task performance and robustness under modality perturbations. Ablation studies confirm the effectiveness of each scheduling factor, and visualization shows that DMS meaningfully adapts to input quality.
Our work highlights the importance of instance-aware and quality-sensitive fusion in building next-generation multimodal systems. As future work, we plan to explore the extension of DMS to more diverse modalities (e.g., audio, video), its application in multi-agent multimodal settings, and its integration with instruction-following and tool-augmented multimodal agents.

\clearpage

{\small
\bibliographystyle{unsrt}
\bibliography{egbib}
}

\end{document}